\documentclass[preprint,5p]{elsarticle}

\usepackage{amsmath,amssymb,amsfonts}
\usepackage{algpseudocode,algorithm}
\usepackage{subfigure}
\usepackage{graphicx}
\usepackage{url}

\journal{Robotics and Autonomous Systems}

\begin{document}

\sloppy

\begin{frontmatter}



\title{Adaptive and Multiple Time-scale Eligibility Traces \\for Online Deep Reinforcement Learning}


\author{Taisuke Kobayashi\corref{cor}}
\ead{kobayashi@is.naist.jp}
\ead[url]{http://kbys\_t.gitlab.io/en/}

\cortext[cor]{Corresponding author}
\address{Nara Institute of Science and Technology, Nara, Japan}

\begin{abstract}
Deep reinforcement learning (DRL) is one promising approach to teaching robots to perform complex tasks.
Because methods that directly reuse the stored experience data cannot follow the change of the environment in robotic problems with a time-varying environment, online DRL is required.
The eligibility traces method is well known as an online learning technique for improving sample efficiency in traditional reinforcement learning with linear regressors rather than DRL.
The dependency between parameters of deep neural networks would destroy the eligibility traces, which is why they are not integrated with DRL.
Although replacing the gradient with the most influential one rather than accumulating the gradients as the eligibility traces can alleviate this problem, the replacing operation reduces the number of reuses of previous experiences.
To address these issues, this study proposes a new eligibility traces method that can be used even in DRL while maintaining high sample efficiency.
When the accumulated gradients differ from those computed using the latest parameters, the proposed method takes into account the divergence between the past and latest parameters to adaptively decay the eligibility traces.
Bregman divergences between outputs computed by the past and latest parameters are exploited due to the infeasible computational cost of the divergence between the past and latest parameters.
In addition, a generalized method with multiple time-scale traces is designed for the first time.
This design allows for the replacement of the most influential adaptively accumulated (decayed) eligibility traces.
The proposed method outperformed conventional methods in terms of learning speed and task quality by the learned policy on benchmark tasks on a dynamic robotic simulator.
A real-robot demonstration confirmed the significance of online DRL as well as the adaptability of the proposed method to a changing environment.
\end{abstract}




\end{frontmatter}


\section{Introduction}

As robotics has advanced significantly over the last few decades, the difficulty of tasks that autonomous robots are expected to complete has increased.
In particular, the tasks with complicated and unknown models (such as human assistance~\cite{modares2015optimized} and cloth manipulation~\cite{tsurumine2019deep}) are expected targets, although they are difficult to be resolved only by classic control methods.
Reinforcement learning (RL)~\cite{sutton2018reinforcement} and its extension combined with deep neural networks (DNNs)~\cite{krizhevsky2012imagenet} to approximate policy and value functions, named deep reinforcement learning (DRL)~\cite{silver2016mastering,levine2018learning}, would be the promising methodology for them.

\begin{figure}[tb]
    \centering
    \includegraphics[keepaspectratio=true,width=0.98\linewidth]{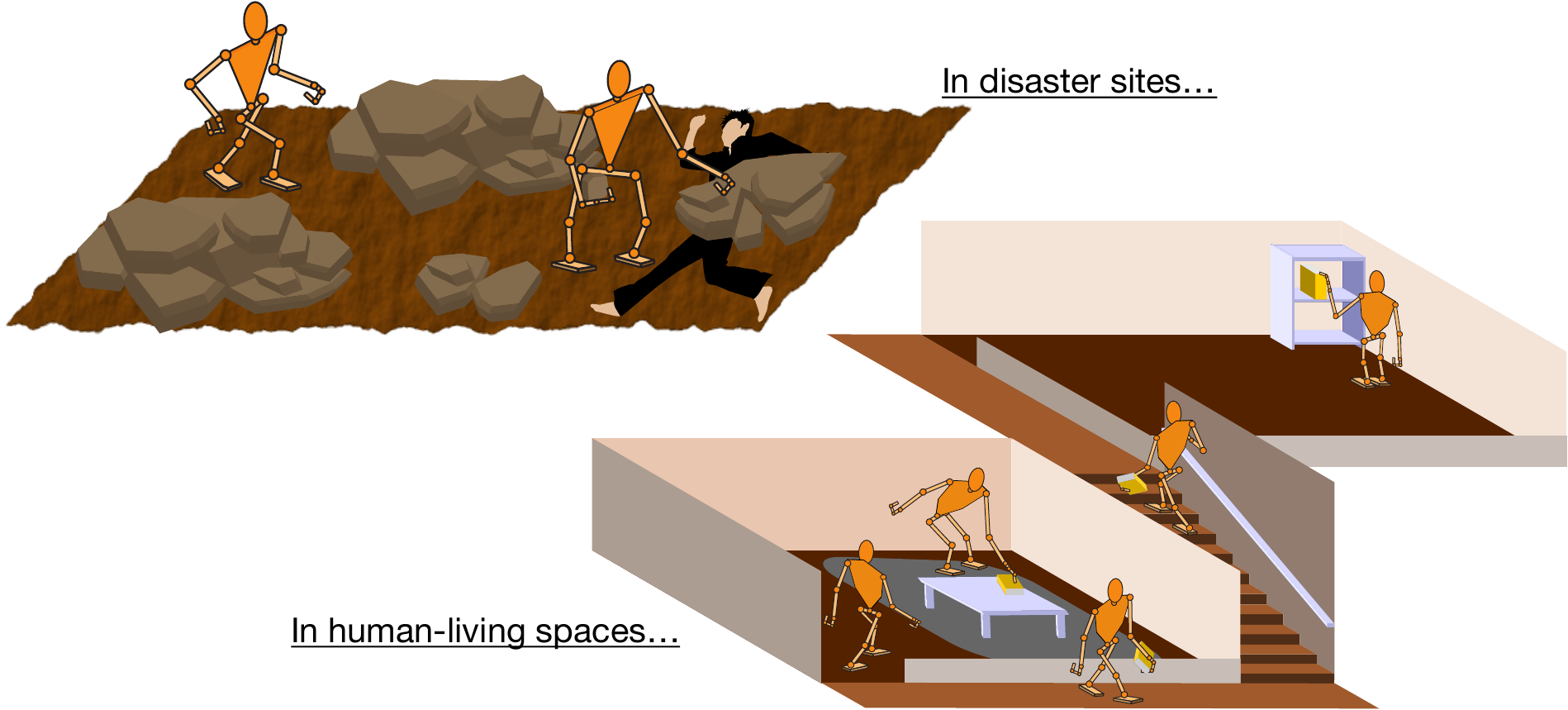}
    \caption{Open problem for autonomous robot learning:
In real-world scenarios, autonomous robots would encounter a wide range of situations with corresponding tasks that are sometimes assigned on the fly, rendering big data collection impractical.
    }
    \label{fig:concept_ARL}
\end{figure}

As an open problem in robotics with DRL, we have to consider its sample efficiency.
In general, RL requires a large number of explorations to find the best task performance from scratch, and learning with DNNs is slower than learning with shallow neural networks due to high nonlinearity and a large number of parameters.
Such sample inefficiency is not a big problem in video games~\cite{silver2016mastering} and stationary tasks for an industry that can collect big data in parallel~\cite{levine2018learning}, but it is infeasible for the autonomous robots.
As illustrated in Fig.~\ref{fig:concept_ARL}, real environments, such as disaster sites and human-living spaces, are non-stationary (or time-varying), and tasks in them are often given on the spot.
Alternatively, abrasion and/or deformation caused by long-term operation may change the kinodynamic characteristics of the robots.
Such situations would prevent collecting sufficient data rapidly.

A naive but practical way to improve the sample efficiency is to store experiences and to replay them later, so-called experience replay~\cite{lin1992self}.
This method can reuse past experiences to update DNNs.
However, in most cases where the autonomous robots have limited computational power and memory capacity, the buffer size to store the experiences would be too much small to achieve good performance~\cite{hayes2019memory}.
Furthermore, as previously stated, autonomous robots must sometimes adapt to non-stationary tasks (e.g., adaptation to human preferences and/or system deterioration over time), and in such cases, previous experiences are no longer reusable.

On the other hand, by fully making use of the properties of RL (i.e., maximizing the sum of rewards in the future and Markovian), the prediction error at the current time is propagated to the past by combining the gradients on the past, so-called eligibility traces~\cite{sutton2018reinforcement,singh1996reinforcement,van2016true}.
This propagation accelerates learning speed by reusing the past gradients.
However, this method is well known as the method only for the traditional RL with linear regressors.
It is reported that DRL with the eligibility traces method tends to fail learning~\cite{van2016effective}, although shallow neural networks can relatively easily be integrated with it~\cite{elfwing2018sigmoid}.

The reason why the eligibility traces fails is explained by Van et al.~\cite{van2016true,van2016effective} that the eligibility traces method is a backward-view approximation of $\lambda$-return~\cite{sutton2018reinforcement,schulman2016high}, which leaves approximation errors.
Although only for linear regressors, such errors have been eliminated by the literature~\cite{van2016true}.
If only that, however, the reason why the standard eligibility traces can utilize with linear regressors is not explained clearly.

Another reason is that the difference between linear and nonlinear regressors is discovered in parameter-dependent gradients.
All of the gradients stored in the eligibility traces are assumed to be derived from the same parameters.
This assumption is met if the gradients are unaffected by parameters, such as the linear regressors.
However, in nonlinear regressors, such as DNNs, the gradients are dependent on other parameters, specifically the parameters of the front layers in DNNs.
That is, the divergence between the gradients computed by the parameters before and after updates (let us call the \textit{gradient divergence}) would be caused as the failure reason.
Although the shallow networks (e.g., only one hidden layer) make the gradient divergence ignorable due to minimal parameter dependency~\cite{elfwing2018sigmoid}, the deep networks have to face it.

The replacing eligibility traces method~\cite{singh1996reinforcement} resets the traces if the latest gradient becomes dominant.
This resetting behavior would mitigate the accumulation of the gradient divergence.
However, because it only uses the most significant past gradient, such a replacing operation is insufficient in terms of sample efficiency.

From the above, two issues that need to be solved in order to use the eligibility traces with DRL are raised:
i) suppression of the gradient divergence for gradients stored in traces,
and ii) improvement of sample efficiency even with replacing operation.
To address these issues, we propose a new eligibility traces method that I decays the eligibility traces adaptively based on gradient divergence, and ii) replaces and retains the most dominant traces rather than the gradient.

Although adaptive decaying is a conservative method, the proposed method can work when recent parameters are only slightly updated.
However, the practical issue is that the gradient divergence is difficult to compute analytically or efficiently.
In practice, therefore, this paper utilizes the Bregman divergences~\cite{bregman1967relaxation} between the past and latest outputs, which imply the gradient divergence and are computed analytically with low computational cost.

The proposed method has multiple time-scale traces for the proposed replacing operation.
Short-term traces of high importance are replaced and stored in long-term traces for an extended period of time.
A hierarchical structure improves sample efficiency while forcing the gradient divergence to be reset by the replacing operation.
Furthermore, this proposed method includes the standard and replacing eligibility traces based on the decaying factor settings (a.k.a. the time scales).
That is, with the right parameters, the proposed method can reap the benefits of both the standard and the replacement eligibility traces.

Four robotic benchmark tasks, including those that either the standard or replacing eligibility traces methods are poor at, are performed in a dynamic simulator to validate the proposed method in terms of general learning performance.
Despite the fact that adaptive decaying has a minor impact on learning performance, the proposed method outperforms conventional methods in all tasks.

\begin{figure}[tb]
    \centering
    \includegraphics[keepaspectratio=true,width=0.97\linewidth]{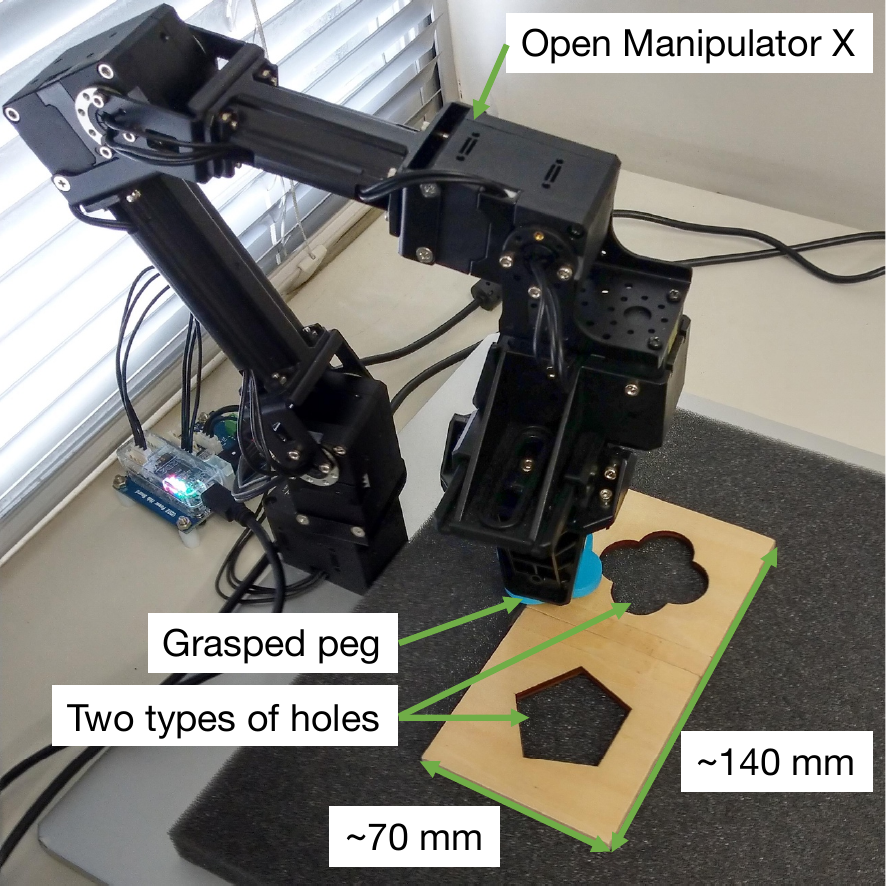}
    \caption{Peg-in-hole task by a four-axis robot arm with a one-axis gripper:
        the task is initialized withholding the peg that fits either of the holes on the left or right, and the robot aims to insert the peg into the hole;
        after learning with one peg, it is exchanged with another to demonstrate adaptability to changes in the problem.
    }
    \label{fig:exp_env}
\end{figure}

Finally, a time-varying peg-in-hole task is demonstrated using a robot arm, Open Manipulator X developed by Robotis (see Fig.~\ref{fig:exp_env}).
Even if the target pair of peg and hole is exchanged (i.e., the reward function changes) after learning for another pair and the previous experience data is no longer reused, the proposed method for online DRL allows the robot to acquire its new target.

\section{Preliminaries}

\subsection{Reinforcement learning}

\subsubsection{Problem statement}

RL makes an agent learn the optimal policy, which can achieve the maximum return (i.e., the sum of rewards) from an environment~\cite{sutton2018reinforcement}.
Here, Markov decision process (MDP) is assumed as the tuple $(\mathcal{S}, \mathcal{A}, \mathcal{R}, p_0, p_T, \gamma)$.
First, the agent gets the initial state $s_0 \in \mathcal{S}$ randomly: $s_0 \sim p_0(s_0)$.
At the time step $t \in \mathbb{N}$, the agent samples an action $a_t \in \mathcal{A}$ from the policy $\pi$ over the current state $s_t \in \mathcal{S}$: $a_t \sim \pi(a_t \mid s_t)$.
The sampled action $a_t$ acts on the environment, and the next state $s_{t+1}$ is sampled according to the transition probability model $p_T$: $s_{t+1} \sim p_T(s_{t+1} \mid s_t, a_t)$.
In addition, the agent gets a reward $r_t \in \mathcal{R}$ according to the reward function: $r_t = r(s_t, a_t, s_{t+1})$.
In this interaction loop, the agent aims to maximize the return defined as $R_t = \sum_{k=0}^\infty \gamma^k r_{t+k}$ where $\gamma \in [0, 1)$ by optimizing the policy to $\pi^*$.
Note that, only if the transition probability model $p_T$ and the reward function $r$ are stationary, the past raw experiences can be reused to enhance the sample efficiency~\cite{lin1992self}.

In many methods, the expected value of the return, named the value function, is approximated as $V(s_t) = \mathbb{E}[R_t \mid s_t]$.
If we have the correct $V$, $\pi^*$ only needs to choose the action that can maximize $V$.
To learn $V$, a temporal difference (TD) error $\delta_t$, which should be minimized, is derived from Bellman equation.
\begin{align}
    \delta_t = r_t + \gamma V(s_{t+1}) - V(s_t)
    \label{eq:td_error}
\end{align}
Note that, to learn the action value function $Q$, We can replace $V(s)$ to $Q(s,a)$.

\subsubsection{Actor-critic algorithm}

Actually, the policy $\pi$ and the value function $V$ are black-box functions, hence it is difficult to learn them directly.
Regression methods are therefore needed to approximate such functions through optimization of their parameters $\theta$.
Note that approximation by DNNs is introduced later.

To learn the optimal parameters, loss functions to be minimized should be designed.
In this paper, two-loss functions for an actor (with the policy), $\mathcal{L}_a$, and a critic (with the value function), $\mathcal{L}_c$, are minimized according to actor-critic algorithms~\cite{schulman2017proximal,haarnoja2018soft,parisi2019td,kobayashi2019student}.
When the policy and value functions are approximated with the parameters $\theta_n$ after $n$ updates, $\pi(a \mid s; \theta_n)$ and $V(s; \theta_n)$, they are defined as follows:
\begin{align}
    \mathcal{L}_a(t, \theta_n) &= - \hat{\delta}_t \log \pi(a_t \mid s_t; \theta_n)
    \label{eq:loss_actor} \\
    \mathcal{L}_c(t, \theta_n) &= - \hat{\delta}_t V(s_t; \theta_n)
    \label{eq:loss_critic}
\end{align}
where $\hat \cdot$ cuts the computational graph and changes the variable to just value.
$\mathcal{L}_a(t, \theta_n)$ is derived from the policy gradient method~\cite{williams1992simple}, and $\mathcal{L}_c(t, \theta_n)$ is equivalent to the minimization of the squared TD error.

In addition, the importance sampling~\cite{tokdar2010importance} is applied.
Although the importance sampling technique was designed for off-policy learning, it can also be used for online on-policy learning without storing the previous parameters, as shown below.
Assume the agent interacts with the environment before updating the policy, and that the policy is updated in parallel with the interaction.
In that case, the action $a_t$ for the interaction has to be sampled from the policy with $\theta_{n-1}$, while $a_t$ is used for updating the policy with $\theta_n$.
\begin{align}
    \rho_{t,n} = \frac{\pi(a_t \mid s_t; \theta_n)}{\pi(a_t \mid s_t; \theta_{n-1})}
    \label{eq:is}
\end{align}

By combining these equations, the main loss function is derived.
\begin{align}
    \mathcal{L}(t,\theta_n) = \hat \rho_{t,n} \left \{ \mathcal{L}_a(t, \theta_n) + \mathcal{L}_c(t, \theta_n) \right \}
    \label{eq:loss_all}
\end{align}
Finally, this loss function is minimized basically using stochastic gradient descent (SGD) methods like Adam~\cite{kingma2014adam}.
Note that the latest regularization techniques~\cite{schulman2017proximal,haarnoja2018soft,parisi2019td} can be integrated into this basic algorithm.

\subsection{Eligibility traces}

\begin{figure}[tb]
    \centering
    \includegraphics[keepaspectratio=true,width=0.98\linewidth]{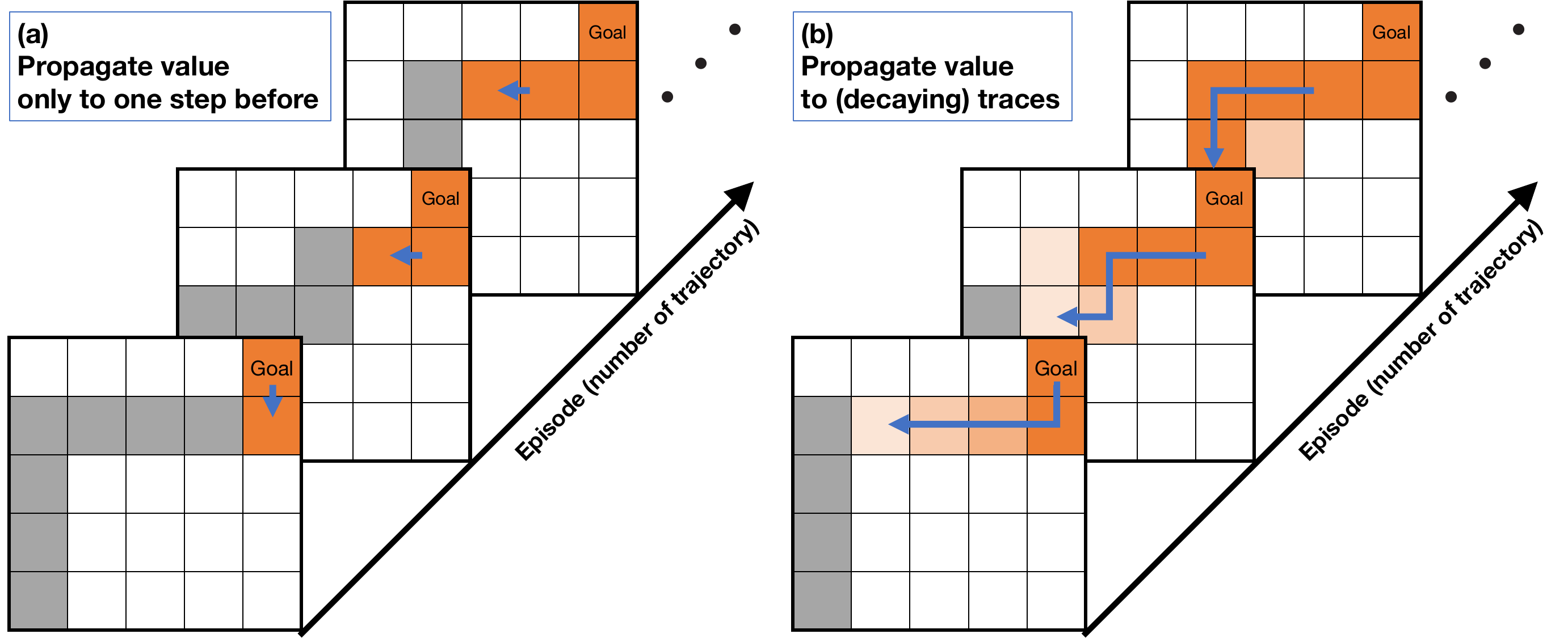}
    \caption{Advantage of eligibility traces:
        in the left~(a), many visits are required to back propagate future information to the initial state;
        in the right~(b), using the eligibility traces, future information can be back propagated effectively.
    }
    \label{fig:eligibility_trace}
\end{figure}

The parameters are updated to minimize the above loss function $\mathcal{L}$ at each time step, but in that case, the propagation of the future rewards to the vicinity of the initial state is too slow, as shown in the left of Fig.~\ref{fig:eligibility_trace}.
To accelerate such propagation, the eligibility traces~\cite{sutton2018reinforcement,singh1996reinforcement,van2016true}, $e$, utilize the trajectory given by MDP, as shown in the right of Fig.~\ref{fig:eligibility_trace}.

That is, the past gradients are accumulated and used to update the parameters as follows:
\begin{align}
    g_t &= \frac{\nabla_{\theta_n} \mathcal{L}(t, \theta_n)}{\hat{\delta}_t}
    \label{eq:grad_loss} \\
    e_t &= \gamma \lambda e_{t-1} + g_t
    \label{eq:eligibility_trace} \\
    \theta_{n+1} &= \theta_n - \alpha \mathrm{SGD}(\hat{\delta}_t e_t)
    \label{eq:sgd}
\end{align}
where $\alpha \in \mathbb{R}_+$ denotes the learning rate and $\mathrm{SGD}(\cdot)$ is one of the SGD methods.
$\lambda \in [0, 1]$ means the rate of decaying.
The eligibility traces are initialized as zero when starting a new trajectory (i.e., at $t = 0$).
Note that, since all the components in $\mathcal{L}$ as shown in eqs.~\eqref{eq:loss_actor} and~\eqref{eq:loss_critic} are multiplied by $\hat{\delta}_t$, this accumulation can be computed stably.

If $\lambda = 0$, this method is consistent with the TD learning without the eligibility traces.
If $\lambda = 1$ and the tasks to be resolved have time limitations, this method represents the Monte Carlo method.
With large $\lambda$, the propagation would be facilitated, thereby improving the sample efficiency.

As a variant of the standard eligibility traces, the replacing eligibility traces~\cite{singh1996reinforcement} have been proposed for the table-style value function.
The original replacing eligibility traces can be extended for the general approximation of the policy and value functions as follows:
\begin{align}
    e_t &=
    \begin{cases}
        g_t & |g_t| > |e_{t-1}|
        \\
        \gamma \lambda e_{t-1} & \mathrm{otherwise}
    \end{cases}
    \label{eq:replacing_trace}
\end{align}
That is, this method stores only the most dominant gradients into $e$.
Although such a replacing operation does not satisfy the backward-view approximation of $\lambda$-return, the better learning performance than by the standard eligibility traces has been reported.

\subsection{DNNs}

When DNNs are employed for approximation of the policy and the value function, $\theta$ is basically given to weight $w$ and biases $b$ for the respective hidden layers and an output layer.
That is, the outputs from the respective hidden layers $x_i$ ($i = 1, \ldots, L$) with $L$ number of layers are defined as follows:
\begin{align}
    x_i = f_i(w_i^\top x_{i-1} + b_i)
\end{align}
where $x_0$ is set as the inputs ($s$ in DRL).
The activation functions, $f_i(\cdot)$, are given for nonlinearity, and they are usually common for all the layers, $f_i(\cdot) = f(\cdot)$.

Finally, the outputs from the last hidden layer is mapped to the domain for the learning targets $y$.
\begin{align}
    y = m(w_o^\top x_L + b_o)
\end{align}
where $m(\cdot)$ denotes the mapping function, such as a softplus function for positive real domain, a sigmoid function for $[0, 1]$, and so on.

\section{Proposed eligibility traces}

\subsection{Problem and solution}

\begin{figure}[tb]
    \centering
    \includegraphics[keepaspectratio=true,width=0.98\linewidth]{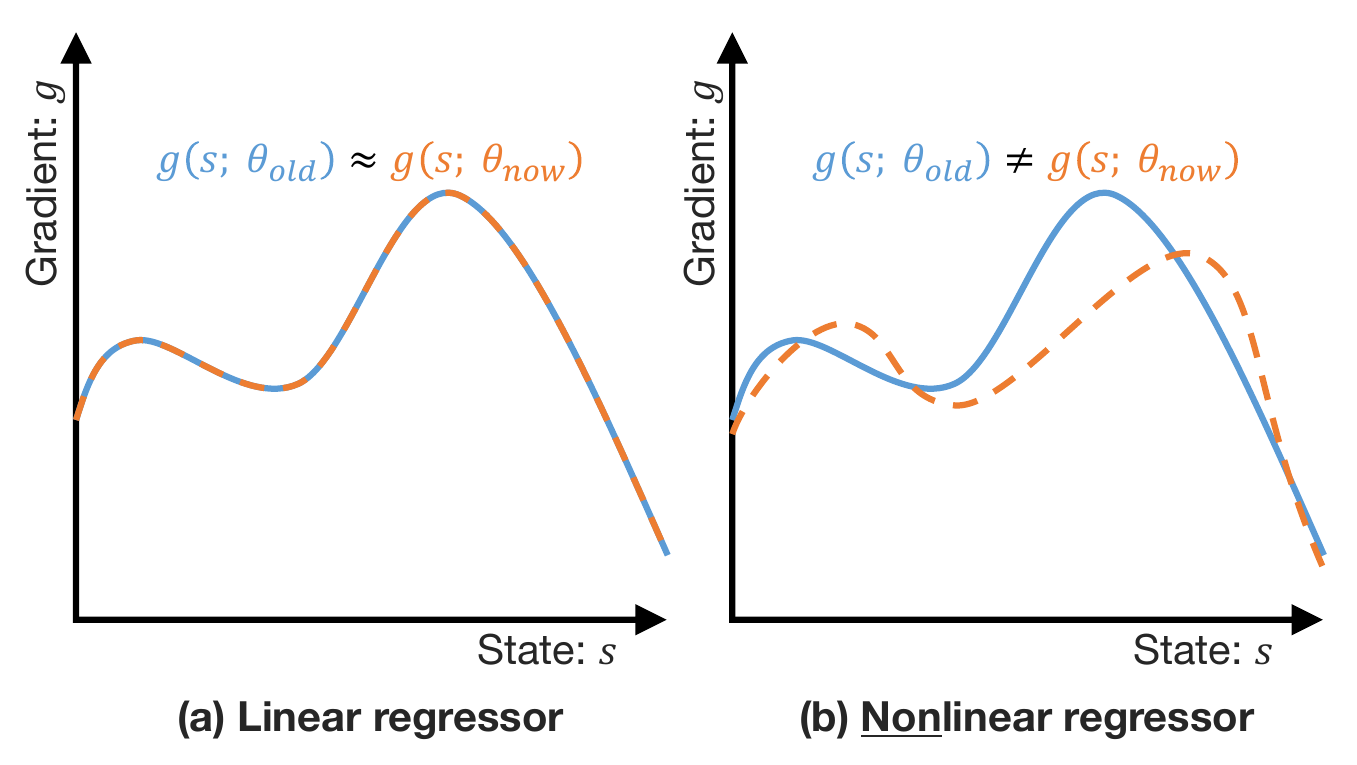}
    \caption{Parameter dependency of gradient in nonlinear regressor:
        unlike the linear regressor shown in the left (a), the nonlinear regressor has the different gradients between the past and latest parameters, as shown in the right (b).
    }
    \label{fig:gradient_divergence}
\end{figure}

Let us theoretically derive the problem hidden in the eligibility traces with the nonlinear regressor (see Fig.~\ref{fig:gradient_divergence}).
Then, a basic idea of its solution is proposed (details are introduced in the next section).

\subsubsection{Case of linear regressor}

The eligibility traces method accumulates the gradients w.r.t the parameters and uses them again and again until they are completely decayed.
If the linear regressors, e.g., $\theta = w$ and $y = m(w^\top x(s))$ with $x(\cdot)$ the fixed function that maps to higher dimensions, the gradients of $y$ w.r.t $w$ are given as follows:
\begin{align}
    \nabla_{\theta_n} y = m^\prime(w_n^\top x(s)) x(s)
    \label{eq:grad_linear}
\end{align}
where $m^\prime(\cdot)$ denotes the derivative of the mapping function $m(\cdot)$.
In addition, the gradients after updating the parameters by $\Delta \theta_n = \theta_{n+1} - \theta_n \ll 1$ is derived as follows:
\begin{align}
    \nabla_{\theta_{n+1}} y = m^\prime(w_n^\top x(s) + \Delta w_n^\top x(s)) x(s)
\end{align}
If the outputs are real by $m(x) = x$ (and $m^\prime(x) = 1$), the above equation is no longer depending on the parameters.
Even if the outputs are in limited spaces like the positive real one, the effects of parameter update is easy to be ignored since $m^\prime(\cdot)$ can be smooth and smaller than $1$, such properties of which can derive $m^\prime(x + \Delta x) - m^\prime(x) < \Delta x$.

\subsubsection{Case of nonlinear regressor}

In contrast, however, nonlinear regressors like DNNs cannot ignore the effects of parameter updates in the gradients.
Specifically, $x(s)$ in eq.~\eqref{eq:grad_linear} is replaced with $x_i$ ($i = 1, \ldots, L$), which is depending on the parameters on $j < i$ layers: $x_i(s; \theta_{j < i})$.
In addition, $y$ is computed multiplication of the features depending on $\theta_{\neq o}$, $x_L(s; \theta_{\neq o})$, and $\theta_o$.
Such highly parameter-dependent gradients may not be reusable for updating the latest parameters as eligibility traces because they differ significantly from the gradients computed by the latest parameters and violate SGD.
Here, the difference between the gradients computed by the past and latest parameters are defined as the \textit{gradient divergence}, $\Delta g(t, \theta_n) = g(t, \theta_{n+1}) - g(t, \theta_n)$.

\subsubsection{Solution}

\begin{algorithm}[tb]
    \caption{Proposed algorithm}
    \label{alg:proposal}
    \begin{algorithmic}[1]
        \State{$n \gets 0$}
        \State{Initialize $\theta_0$}
        \While{True}
            \State{$t \gets 0$, $e = 0$, and assume $\theta_{n-1} = \theta_n$}
            \Comment{Reset}
            \State{$s_0 \sim p_0(s_0)$}
            \While{True}
                \State{Compute $\pi(s_t; \theta_n), V(s_t; \theta_n)$}
                \State{$a_t \sim \pi(a_t \mid s_t; \theta_n)$}
                \State{Execute $a_t$}
                \If{$t \neq 0$}
                    \Comment{During interaction if possible}
                    \State{Compute the gradient $g_t$}
                    \State{\ \ using eqs.~\eqref{eq:td_error}--\eqref{eq:grad_loss}}
                    \State{Compute the adaptive decaying factor $\lambda^d_{t,n}$}
                    \State{\ \ using eqs~\eqref{eq:div_policy}--\eqref{eq:adaptive_decay}}
                    \State{Update parameters to $\theta_{n+1}$}
                    \State{\ \ using eqs.~\eqref{eq:generalized_trace} and~\eqref{eq:sgd2}}
                    \State{Compute $\pi(s_t; \theta_{n+1}), V(s_t; \theta_{n+1})$}
                    \State{$n \gets n + 1$}
                \EndIf
                \State{$s_{t+1} \sim p_T(s_{t+1} \mid s_t, a_t)$}
                \State{$t \gets t + 1$}
                \If{Meet end conditions}
                    \State{\textbf{break}}
                \EndIf
            \EndWhile
        \EndWhile
    \end{algorithmic}
\end{algorithm}

As previously stated, gradient divergence is one possible reason why the eligibility traces in DNNs fail.
To eliminate this, the naive solution is to recompute the loss functions given by the past states, actions, and rewards (i.e., to directly use $\lambda$-return~\cite{sutton2018reinforcement,van2016effective,schulman2016high}), although that requires high computational cost and memory capacity.
The eligibility traces have the advantages of low computational cost and memory efficiency, making them suitable for online learning in autonomous robots.
As a result, we must modify the eligibility traces in some way to mitigate or avoid the negative effects of gradient divergence.

As a solution, this paper proposes a conservative method of adaptively decaying the eligibility traces with high gradient divergence.
To do so, the gradient divergence should be quantified.
As an alternative to gradient divergence, the following section introduces the definition of output divergences.
The adaptive decaying method is proposed based on the output divergences.

In addition, the resetting behavior of the replacing eligibility traces method~\cite{singh1996reinforcement} would implicitly mitigate the accumulation of the gradient divergence.
In light of this, a generalized eligibility traces method with multiple time-scale traces is proposed, which includes the standard and replacing eligibility traces methods.

In summary, the implemented algorithm is written in Alg.~\ref{alg:proposal}.

\subsection{Output divergences as alternative to gradient divergence}

\begin{figure}[tb]
    \centering
    \includegraphics[keepaspectratio=true,width=0.8\linewidth]{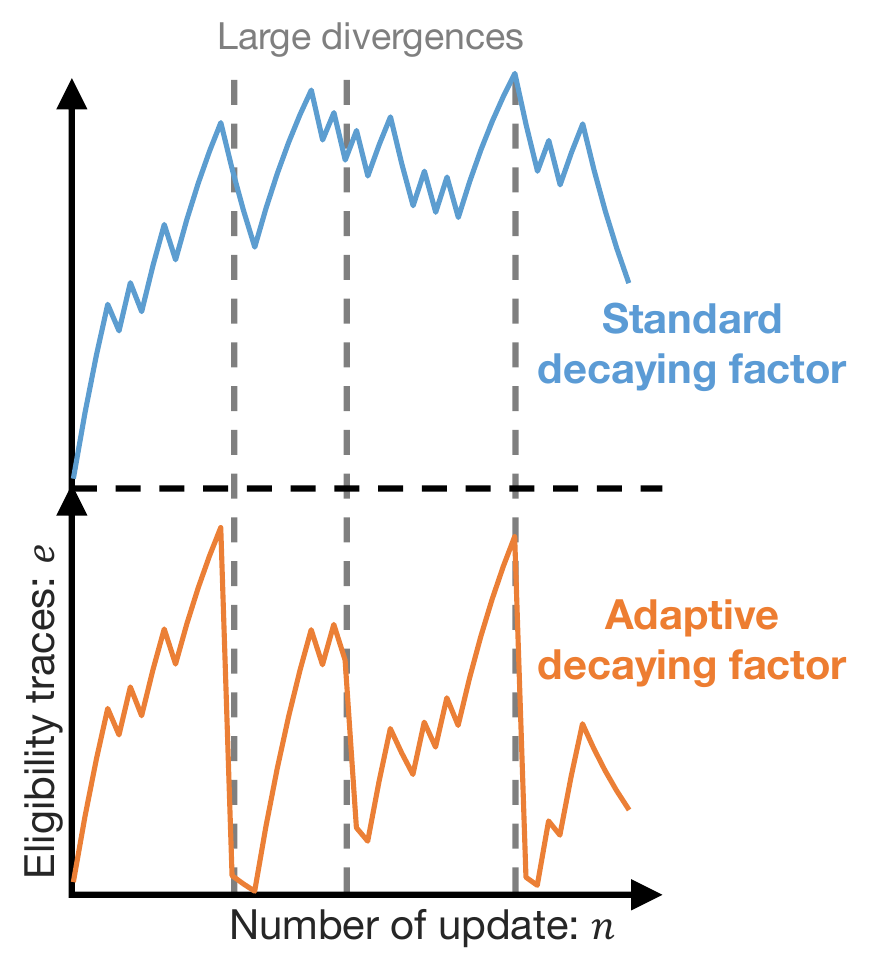}
    \caption{Behavior of the adaptive decaying factor:
        if the large divergences happen, the adaptive decaying resets the eligibility traces since the stored gradients no longer indicate the correct update direction.
    }
    \label{fig:adaptive_decaying_factor}
\end{figure}

As previously stated, the gradient divergence is difficult to define in a computationally cheap manner due to the large parameter space.
Instead, we focus on that the outputs of regressors, i.e., the policy $\pi$ and the value function $V$, are in relatively small spaces and reveal the change in the parameters.
That is, the gradient divergence is approximated as the combination of two output divergences of $\pi$ and $V$.

By considering the types of the respective spaces, the divergences are generally given as Bregman divergence~\cite{bregman1967relaxation}.
Specifically, the divergence of $\pi$, which is in probability space, is given as Kullback-Leibler (KL) divergence.
\begin{align}
    d^\pi_{t,n} &= \mathrm{KL}(\pi(a \mid s_t; \theta_n) \mid \pi(a \mid s_t; \theta_{n-1}))
    \nonumber \\
    &= \int_a \pi(a \mid s_t; \theta_n) \log \rho_{t,n}(a) da
    \label{eq:div_policy}
\end{align}
Note that, in most cases, this integral can be solved with a closed-form solution for the model that is employed for the policy (e.g., normal distribution).
In addition, the divergence of $V$, which is in Euclid space, is given as L2 norm.
\begin{align}
    d^V_{t,n} = \frac{1}{2}\left \{ V(s_t; \theta_n) - V(s_t; \theta_{n-1}) \right \}^2
    \label{eq:div_value}
\end{align}

As a remark, even in the cases without the closed-form solution of KL divergence, Pearson divergence~\cite{pearson1900x} would alternate with it.
\begin{align}
    d^\pi_{t,n} &= \int_a \pi(a \mid s_t; \theta_{n-1}) (\rho_{t,n}(a) - 1)^2 da
    \nonumber
\end{align}
Because this definition is based on a squared loss, the Pearson divergence is always positive, even when the Monte Carlo approximation is used, whereas the KL divergence is not.

Using the above two divergences, $d^\pi$ and $d^V$, $\lambda$ for adaptively decaying the eligibility traces are defined as follows:
\begin{align}
    d^s_{t,n} &= \lambda^d_{t,n-1} d^s_{t,n-1} + \left (d^\pi_{t,n} + d^V_{t,n} \right)
    \label{eq:div_sum} \\
    \lambda^d_{t,n} &= \exp \left (- \kappa d^s_{t,n} \right)
    \label{eq:adaptive_decay} \\
    \lambda_{t,n} &= \lambda_\mathrm{max} \lambda^d_{t,n}
\end{align}
where $\lambda_\mathrm{max}$ is the maximum of the decaying factor since $d^\pi, d^V > 0$, and $\lambda=\lambda_\mathrm{max}$ in the standard method.
$\kappa$ denotes the gain to adjust the ease of decaying.

$d^s$ accumulates the divergences while decaying adaptively.
If $d^\pi$ and $d^V$ are instantaneously large (i.e., the parameters are largely changed), the eligibility traces $e$ are defined in eq.~\eqref{eq:eligibility_trace} are reseted since the accumulated gradients are likely to be different from that computed by the latest parameters.
If $(1-\lambda^d) d^s \simeq d^\pi + d^V$, the decaying speed and the accumulated divergences are balanced, and $\lambda^d$ would converge on the specific value (less than $1$).
That means the decaying factor is adjusted constantly.
Otherwise, the gradients would be accumulated to $e$ like the standard method does (see Fig.~\ref{fig:adaptive_decaying_factor}).

\subsection{Generalization to include replacing eligibility traces}

\begin{figure}[tb]
    \centering
    \includegraphics[keepaspectratio=true,width=0.98\linewidth]{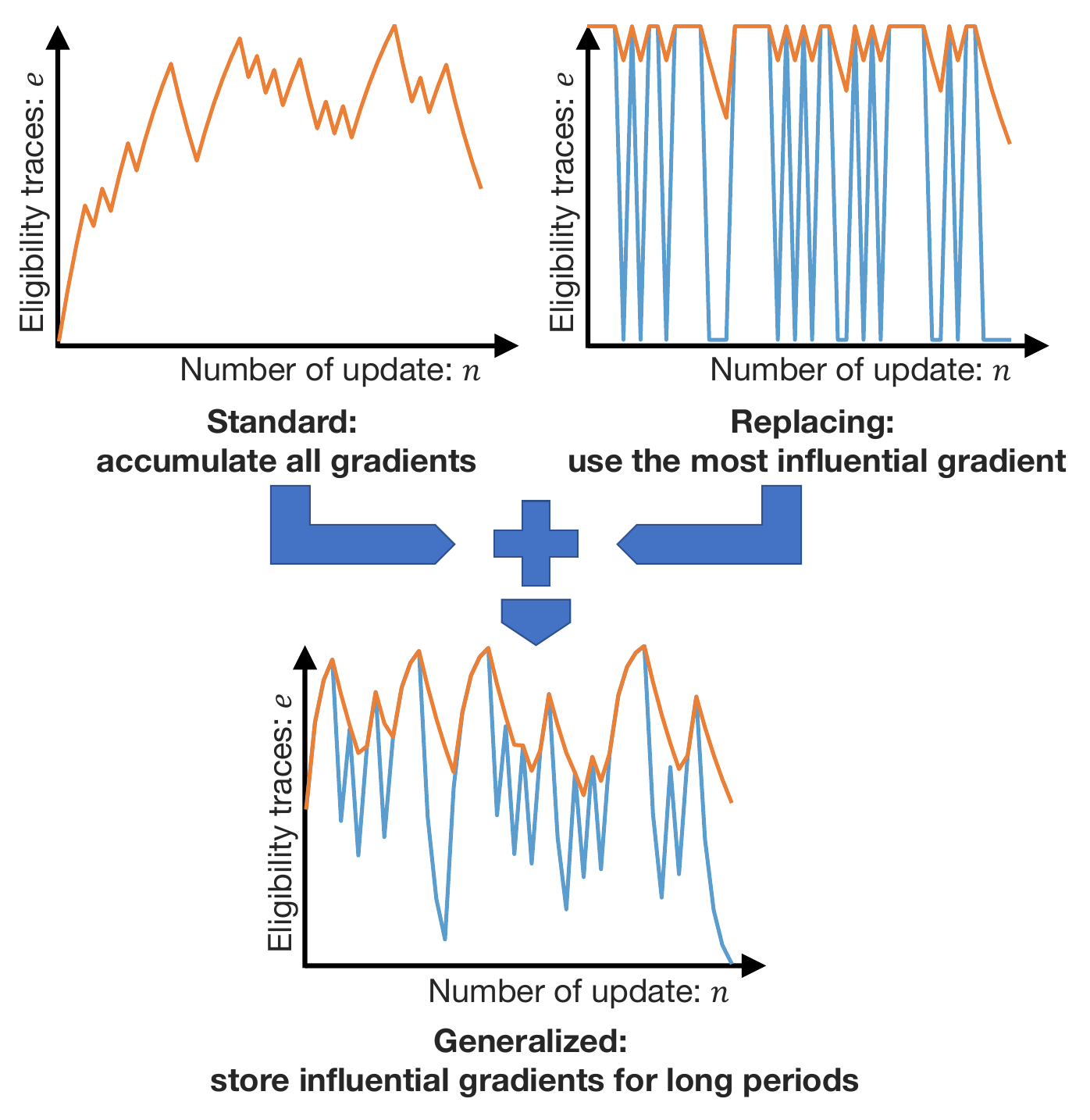}
    \caption{Behavior of the generalized eligibility traces:
        by having two different time-scale traces, this new method can accumulate the gradients as like the standard eligibility traces, and store them for long periods by replacing them into other traces as like the replacing eligibility traces.
    }
    \label{fig:generalized_trace}
\end{figure}

As shown in eq.~\eqref{eq:replacing_trace}, the replacing eligibility traces discard all the past gradients if the new gradient exceeds the stored one.
This discarding is expected to function in the same way as the adaptive decaying described above.
On the other hand, by shortening the trajectory, the replacing eligibility traces would fall short of propagating the most recent value to the past.
Thus, generalized eligibility traces are proposed in this section to take advantage of the benefits of replacing eligibility traces while accumulating trajectory information.

Let us define $e^i$ ($i = 1, \ldots, K$) with $K > 1$ as multiple time-scale traces.
They are with a $K$-series-layered structure, and the maximum decaying factor of $i$-th layer, $\lambda_\mathrm{max}^i$, is given.
Under these conditions, a new update rule to replace conventional ones (i.e., eqs~\eqref{eq:eligibility_trace} and~\eqref{eq:sgd} for the standard version and; eqs~\eqref{eq:replacing_trace} and~\eqref{eq:sgd} for the replacing version) is proposed as follows:
\begin{align}
    e_t^i &=
    \begin{cases}
        e_t^{i-1} & i \neq 1 \land \Delta e^i e_t^{i-1} > 0
        \\
        \gamma \lambda_\mathrm{max}^i \lambda^d_{t,n} e_{t-1}^i + \beta^i g_t & \mathrm{otherwise}
    \end{cases}
    \label{eq:generalized_trace} \\
    \theta_{n+1} &= \theta_n - \alpha \mathrm{SGD}(\hat{\delta}_t e_t^K)
    \label{eq:sgd2}
\end{align}
where $\Delta e^i = e_t^{i-1} - e_{t-1}^i$.
$\beta^i$ is the sequence with three conditions: $\beta^i \geq \beta^{i+1}$, $\sum_{i=1}^K \beta^i = 1$, and $\beta^K = 0$.
For example, the following arithmetic sequence satisfies the above conditions:
\begin{align}
    \beta^i = \frac{2 (K - i)}{K (K - i)}
    \label{eq:eg_beta}
\end{align}

This update rule implies that the gradient propagates from shallow to deep layers with different decay time constants.
It is worth noting that the condition for replacing has been relaxed in order to make the sign reversal between layers true.
This kind of relaxation would allow the update direction to quickly follow the most recent gradient direction.

The large $K$ wastes system memory capacity and negates the benefits of eligibility traces in comparison to experience replay.
Therefore, $K$ is fixed on two in this paper, which is the minimum value to gain the benefits from replacing eligibility traces.
In fact, although the above equations are given in a general form, it is easy to imagine that the effects of deepening the layer are not as critical due to the effects of adaptive decaying, which adaptively resets all the traces.

In the case with $K = 2$, the proposed eligibility traces can generalize the standard/replacing eligibility traces.
Specifically, we expect three modes according to the setting of $\lambda_\mathrm{max}^1$ and $\lambda_\mathrm{max}^2$ as follows (also, see Fig.~\ref{fig:generalized_trace}):
\begin{enumerate}
    \item $\lambda_\mathrm{max}^1 > \lambda_\mathrm{max}^2$:
    The second layer decays its traces faster than the first layer, that is, we can expect $e_t^1 \simeq e_t^2$.
    Specifically, if all the gradients are positive (or negative) or $\lambda_\mathrm{max}^2 = 0$, this mode perfectly matches the standard eligibility traces.
    \item $\lambda_\mathrm{max}^1 = 0$ and $\lambda_\mathrm{max}^2 > 0$:
    Since $e_t^1 = g_t$, this mode is almost the same as the replacing eligibility traces.
    The difference is only whether the sign reversal is included in the replacing condition.
    \item $0 < \lambda_\mathrm{max}^1 < \lambda_\mathrm{max}^2$:
    The traces with the short-term memory $e^1$ replace the traces with the long-term memory $e^2$ only if $e^1$ may be more influential than $e^2$.
    As a result, only the most important traces are actively stored for long periods.
    Such a behavior can be regarded to be similar to hyperbolic discounting, which has been studied as a biologically plausible decision-making mechanism with variable decaying speeds~\cite{rachlin1972commitment,kobayashi2008influence}.
\end{enumerate}
From the above analysis, the third mode should be better than the others.
To exploit the advantages of both the standard and replacing eligibility traces, it is expected that $\lambda_\mathrm{max}^2 \simeq 2 \lambda_\mathrm{max}^1$ would be a reasonable choice.

\section{Simulations}

\begin{table*}[tb]
    \caption{Simulation environments provided by Pybullet Gym~\cite{brockman2016openai,coumans2016pybullet}}
    \label{tab:sim_environment}
    \centering
    \begin{tabular}{ccccc}
        \hline\hline
        ID & Name & State space $|\mathcal{S}|$ & Action space $|\mathcal{A}|$ & Episode $E$ \\
        \hline
        InvertedPendulumBulletEnv-v0 & InvertedPendulum & 5 & 1 & 200 \\
        InvertedPendulumSwingupBulletEnv-v0 & Swingup & 5 & 1 & 200 \\
        HalfCheetahBulletEnv-v0 & HalfCheetah & 26 & 6 & 2000 \\
        AntBulletEnv-v0 & Ant & 28 & 8 & 2000 \\
        \hline\hline
    \end{tabular}
\end{table*}

\begin{figure*}[tb]
    \centering
    \includegraphics[keepaspectratio=true,width=0.98\linewidth]{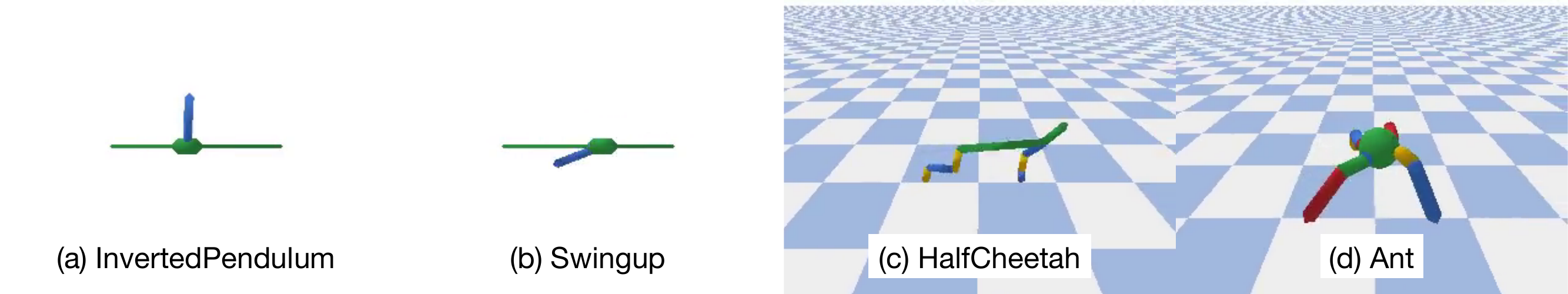}
    \caption{Snapshots of four benchmark tasks:
        they are simulated by Pybullet Gym~\cite{brockman2016openai,coumans2016pybullet};
        (a) InvertedPendulum makes a cart keep a pole standing;
        (b) Swingup makes a cart swing up a pole on it standing;
        (c) HalfCheetah makes a two-legged robot walk forward as fast as possible;
        (d) Ant makes a four-legged robot walk forward as fast as possible.
    }
    \label{fig:sim_env_snap}
\end{figure*}

\subsection{Benchmark tasks}

To begin, the proposed method's performance as an RL method is compared to conventional methods in general RL benchmark simulations.
Because these general RL benchmarks are stationary problems, the adaptability of the proposed method to non-stationary problems, which is one of its advantages, is not evaluated here, but will be validated in real robot experiments in the following section.

In the dynamic simulations, the agent learns the given tasks, which have $|\mathcal{S}|$-dimensional state space and $|\mathcal{A}|$-dimensional action space, through $E$ episodes with up to $T = 1000$ time steps.
After learning, the agent repeats the learned task 50 times to compute the sum of rewards for each, and the median of these is used to calculate the score.
The agent attempts to learn each task 20 times with different random seeds for statistical evaluation.

Four benchmark tasks for DRL simulated by Pybullet Gym~\cite{brockman2016openai,coumans2016pybullet} are prepared.
They are listed in Table~\ref{tab:sim_environment} (also see Fig.~\ref{fig:sim_env_snap}).
Their purposes (i.e., their reward designs) are given as follows:
\begin{enumerate}
    \renewcommand{\labelenumi}{(\alph{enumi})}
    \item InvertedPendulum:
    A cart keeps a pole standing.
    \item Swingup:
    A cart swings up a pole and keeps it standing.
    \item HalfCheetah:
    A two-dimensional cheetah with two legs walks forward as fast as possible.
    \item Ant:
    A three-dimensional quadruped walks forward as fast as possible.
\end{enumerate}

\subsection{Conditions}

\begin{figure}[tb]
    \centering
    \includegraphics[keepaspectratio=true,width=0.98\linewidth]{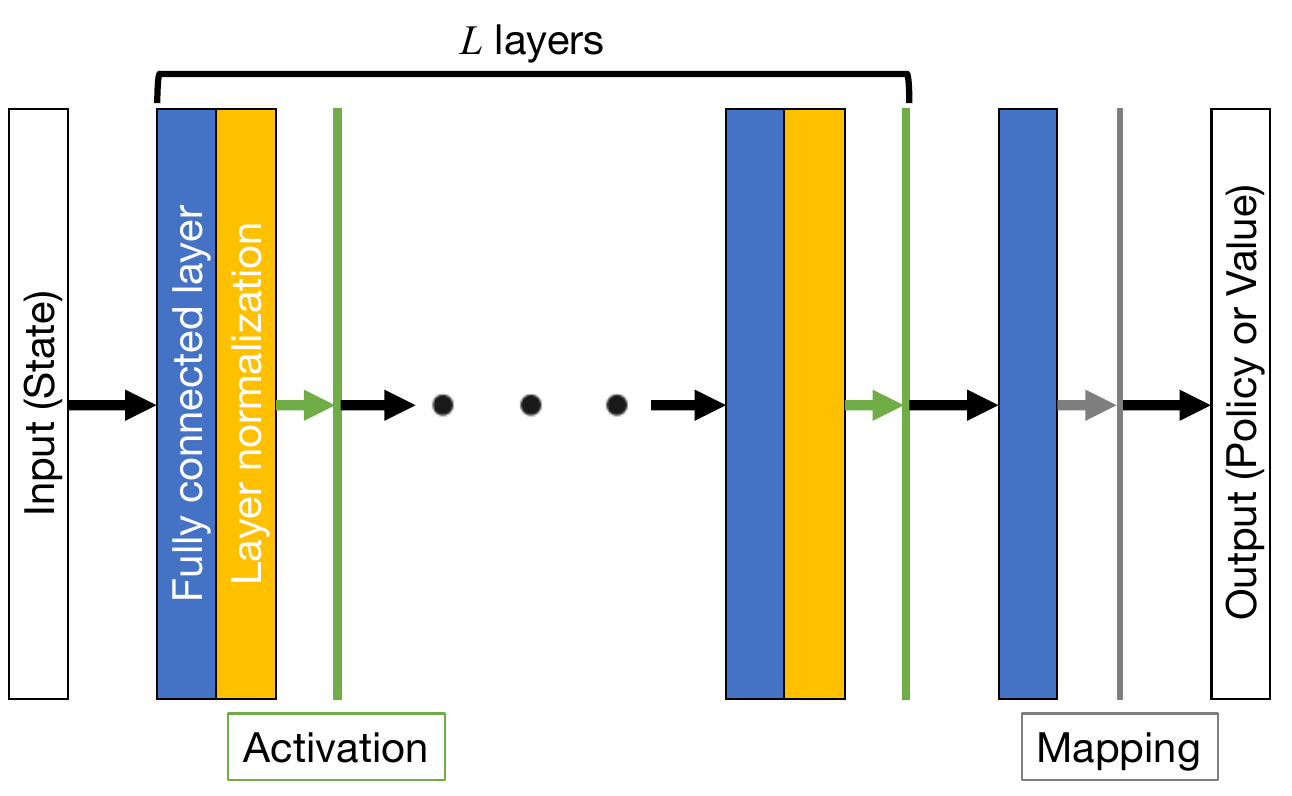}
    \caption{Network architecture in actor for policy or critic for value function:
        the input is given to $L$ modules, which are connected in series;
        each module is composed of a fully connected layer, a layer normalization, and an activation function;
        an additional fully connected layer uses for shaping the features given by $L$ modules to the output.
    }
    \label{fig:network_architecture}
\end{figure}

Basic network architecture, which is implemented by PyTorch~\cite{paszke2017automatic}, is shown in Fig.~\ref{fig:network_architecture}.
The tasks conducted are without an image input state.
Therefore, only $L$ fully connected layers with $N$ neurons are employed.
To avoid overfitting, layer normalization~\cite{ba2016layer} is added after each layer except the output one.
As an activation function, the Swish function~\cite{ramachandran2017swish,elfwing2018sigmoid} is employed for biological plausibility and high nonlinearity.

Using this network architecture, the actor and critic approximate the policy $\pi$ and value function $V$, respectively.
Specifically, as the policy model, student-t distribution~\cite{kobayashi2019student}, which has location parameter $\mu \in \mathbb{R}^{|\mathcal{A}|}$, scale parameter $\sigma \in \mathbb{R}^{|\mathcal{A}|}_+$, and degrees of freedom $\nu \in \mathbb{R}_{\geq 2}$ as parameters to be approximated, is employed.
Since $\sigma$ and $\nu$ are in the positive real domain, the soft plus function is used as the mapping function.
The others (i.e., $\mu$ and $V$) are in the real domain, so an identity map is used.

To stably learn the tasks, the latest policy regularization techniques are combined.
Specifically, the importance sampling is replaced by PPO~\cite{schulman2017proximal} with a clipping value $\epsilon$.
The policy entropy regularization based on SAC~\cite{haarnoja2018soft} is added with a regularization weight $\beta_{DE}$.
The TD regularization~\cite{parisi2019td} is also introduced with a regularization weight $\beta_{TD}$.
In addition, for robustness to noisy gradients in online learning, a robust SGD, i.e., LaProp~\cite{ziyin2020laprop} with t-momentum~\cite{ilboudo2020robust} and d-AmsGrad~\cite{kobayashi2021towards} (so-called td-AmsProp), is employed with their default parameters except the learning rate.
It should be noted that the above implementation was obtained through trial and error because online DRL failed to complete the most recent RL benchmark tasks.

Table~\ref{tab:sim_parameter} summarizes the common hyperparameters to be used in the simulations.
Here, these parameters were tuned empirically using CartPoleContinuousBulletEnv-v0.
In addition, the hyperparameters related to the proposed method, $(\lambda_\mathrm{max}^1, \lambda_\mathrm{max}^2, \kappa)$, are set as the following comparative conditions.
\begin{enumerate}
    \item $(0.0, 0.0, 0.0)$ as no eligibility traces
    \\(labeled ``No'')
    \item $(0.9, 0.0, 0.0)$ as standard eligibility traces
    \\(labeled ``Standard'')
    \item $(0.0, 0.9, 0.0)$ as replacing eligibility traces
    \\(labeled ``Replacing'')
    \item $(0.9, 0.0, 1.0)$ as adaptive standard eligibility traces
    \\(labeled ``Adapt+Standard'')
    \item $(0.0, 0.9, 1.0)$ as adaptive replacing eligibility traces
    \\(labeled ``Adapt+Replacing'')
    \item $(0.5, 0.9, 1.0)$ as proposed eligibility traces
    \\(labeled ``Proposed'')
\end{enumerate}
Here, $\lambda_\mathrm{max} = 0.9$ is given based on the preferred value for the standard eligibility traces.
Through tuning, it is empirically found that the better $\kappa$ is within $[1.0, 10.0]$.
The source codes with the above implementation can be downloaded from \url{https://github.com/kbys-t/test_et.git}.

Because the proposed method includes the conventional eligibility traces, it can be compared to the conventional methods as an ablation test by using the hyperparameters described above.
The significance of the eligibility traces will be confirmed in particular by comparing the first condition to the others.
The adaptive decaying will be validated by comparing standard/replacing eligibility traces with and without them.
By comparing the last three conditions, the proposed eligibility traces with multiple time-scale traces will also be verified.

\begin{table}[tb]
    \caption{Common hyperparameters for the simulations}
    \label{tab:sim_parameter}
    \centering
    \begin{tabular}{ccc}
        \hline\hline
        Symbol & Meaning & Value \\
        \hline
        $N$ & Number of neurons & 128 \\
        $L$ & Number of layers & 5 \\
        $\gamma$ & Discount factor & 0.99 \\
        $\alpha$ & Learning rate & 1e-4 \\
        $\epsilon$ & Threshold for clipping & 0.1 \\
        $\beta_{DE}$ & Gain for entropy regularization & 0.025 \\
        $\beta_{TD}$ & Gain for TD regularization & 0.025 \\
        \hline\hline
    \end{tabular}
\end{table}

\subsection{Results}

\begin{figure}[tb]
    \centering
    \includegraphics[keepaspectratio=true,width=0.98\linewidth]{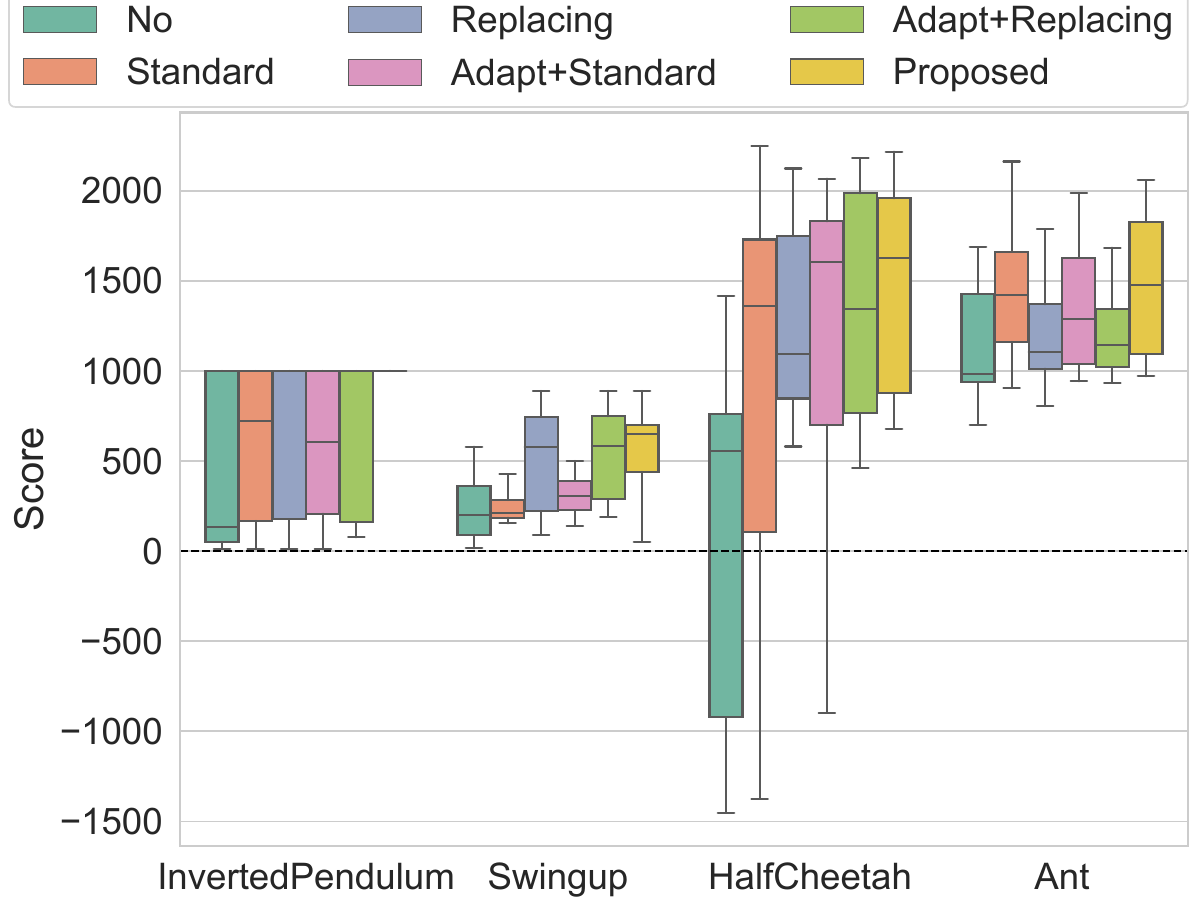}
    \caption{Test results of four benchmark tasks illustrated as box plots:
        although the contribution of the adaptive decaying was not significant, we found improvements at the median level;
        thanks to the integration of the standard and replacing eligibility traces, the proposed method (with yellow) outperformed the others in all the tasks.
    }
    \label{fig:sim_summary}
\end{figure}

\begin{figure*}[tb]
    \centering
    \includegraphics[keepaspectratio=true,width=0.99\linewidth]{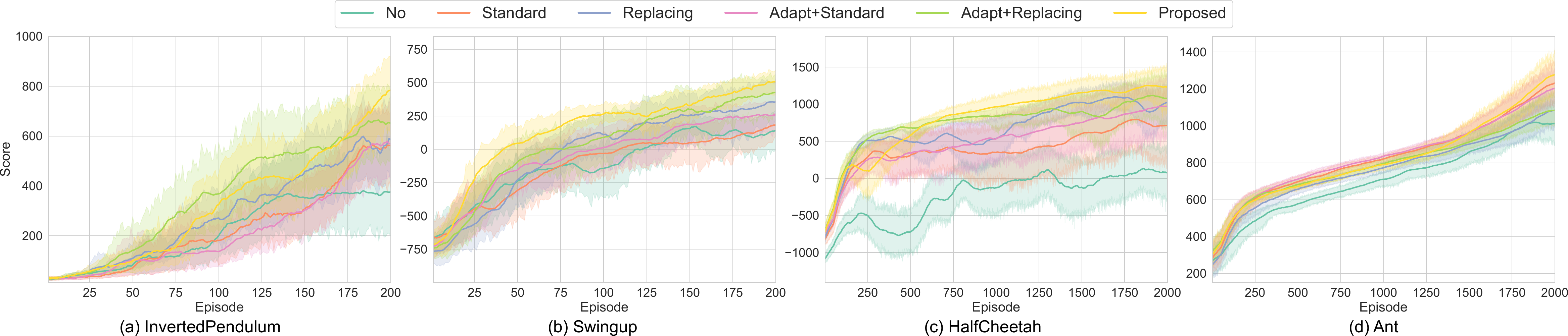}
    \caption{Learning curves of four benchmark tasks:
        the sum of rewards at each episode is given as the score;
        the corresponding shaded areas show the 95~\% confidence intervals;
        the proposed method (with yellow) increased the return rapidly, and outperformed the others in total.
    }
    \label{fig:sim_learn}
\end{figure*}

\begin{figure*}[tb]
    \centering
    \includegraphics[keepaspectratio=true,width=0.99\linewidth]{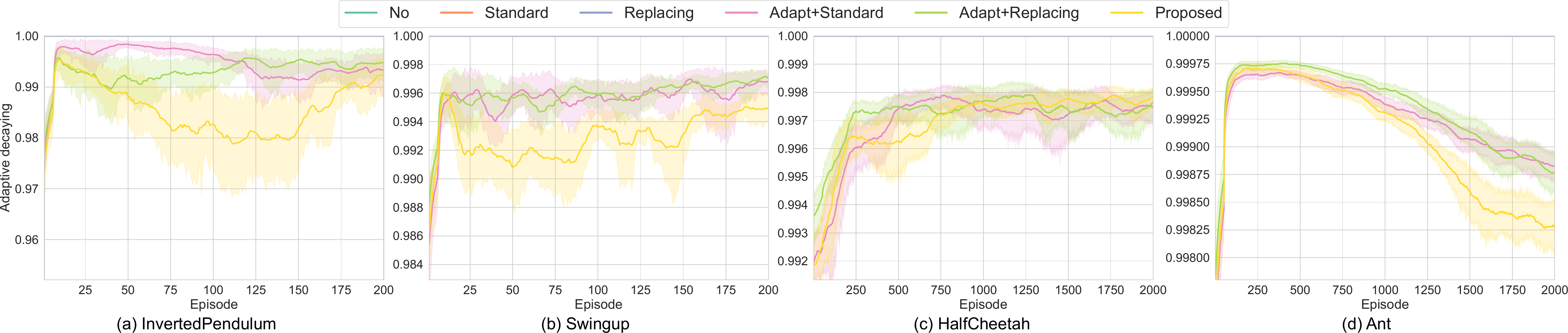}
    \caption{Average $\lambda^d$ of four benchmark tasks:
        the corresponding shaded areas show the 95~\% confidence intervals;
        the proposed method (with yellow) decayed the eligibility traces more frequently than the cases with $(0.9, 0.0, 1.0)$ and $(0.0, 0.9, 1.0)$;
        the frequent decaying implies that the parameters $\theta$ were actively updated and the optimization progressed appropriately.
    }
    \label{fig:sim_decay}
\end{figure*}

First, to confirm the learning tendency, the test results with the respective scores (i.e., the sum of rewards at each episode) are summarized in Fig.~\ref{fig:sim_summary}.
Overall, the DRL failed to learn all of the tasks without the eligibility traces.
In contrast, the eligibility traces allowed the agent to obtain the assigned tasks.
The proposed decaying technique in eqs.~\eqref{eq:div_policy}--\eqref{eq:adaptive_decay} did not improve the performance of the tasks significantly, although the medians indicated as the lines in the boxes were likely to be increased by the adaptive decaying with $\kappa = 1$.
The reason why the standard eligibility traces without the adaptive decaying succeeded in learning more than expected would be that the latest regularization techniques~\cite{schulman2017proximal,haarnoja2018soft,parisi2019td} suppress the amount of parameter update, and secondarily reduce the gradient divergence, which is considered as the reason why the eligibility traces are unsuitable for DRL.

Remarkably, the proposed method with eqs.~\eqref{eq:generalized_trace} and~\eqref{eq:sgd2} outperformed the others in all the tasks, as can be seen in the yellow boxes in Fig.~\ref{fig:sim_summary}.
Whereas the standard/replacing eligibility traces have task-specific performances, particularly in locomotion tasks (HalfCheetah and Ant tasks), the proposed eligibility traces performed well in both tasks, owing to the integrated characteristics.
As a result, we can agree that both proposed techniques for eligibility traces (particularly the generalized version with multiple time-scale traces) significantly improved learning performance.

The learning behaviors of the score are shown in Fig.~\ref{fig:sim_learn}.
In addition, the average of the adaptive decaying factors $\lambda^d$ in each episode is depicted in Fig.~\ref{fig:sim_decay}.
We found that the decaying tends to be strengthened when the score changes greatly (in particular, increases).
For example, in InvertedPendulum, the score increased around 75--125 episodes as the task started to be successful, and $\lambda^d$ at that time was relatively smaller than other episodes.
Swingup had the same relationship with HalfCheetah around 25--50 episodes, HalfCheetah at the start of 250 episodes, and Ant after 1500 episodes.
This is a natural behavior because when performance improves, the policy and value functions are greatly updated.
In fact, the decaying was stronger overall in the proposed method with the highest performance $(0.5, 0.9, 1.0)$ than in the others.
However, as previously stated, the proposed decaying did not significantly improve task performance; thus, it is desired to investigate the value of such adaptive decaying and improve its design in future work.

By focusing on the learning curves in Fig.~\ref{fig:sim_learn}, we also found the sample efficiency of the proposed method.
We can see that the method with no eligibility traces had a lower score than the others at the start of the study, implying that the eligibility traces accelerated learning.
When the proposed method was compared to the methods with the conventional (standard/replacing) eligibility traces, the proposed method outperformed the conventional methods in general.
From the start, the proposed method outperformed the conventional methods in Swingup.
In Ant, the proposed method's learning was relatively delayed compared to the standard eligibility traces method, but it quickly rose in the latter half of the learning period and outperformed the others.
This implies that the proposed method was capable of escaping the situation, in which it was stuck in the local solution by appropriately accumulating and reusing gradients with small gradient divergence.
From these results, we can conclude that the proposed method improved the sample efficiency without reusing the past experiences directly, resulting in the highest score shown in Fig.~\ref{fig:sim_summary}.

\section{Demonstration with real robot experiments}

\subsection{Peg-in-hole task}

Because the proposed method's learning performance as a general RL method has been confirmed in the preceding simulations, its adaptability to non-stationary problems is demonstrated.
As an example of a non-stationary problem, the change in reward is used, as it is easy to see the different optimal results.
It should be noted that the following task is intended to be a case-study demonstration of qualitative adaptability to non-stationary problems rather than a quantitative comparative evaluation with RL methods.

Specifically, as a real-robot-used demonstration of the proposed method, a peg-in-hole task with a robot arm is performed (see Fig.~\ref{fig:exp_env}).
The robot used is Open Manipulator X developed by Robotis, which has one axis for yaw rotation; three axes for pitch rotation; and one axis for a gripper.
That is, the robot can control the position of its end-effector while keeping the gripper vertical.
However, it should be noted that the position control is not sufficiently accurate and the control error may occur.

Two different types of pegs are prepared, and either of them is grasped by the robot from the start of the episode.
Two 70~mm squared plates, each of which has the hole corresponding to either of the pegs in the center, are set side by side on the environment.
The depth of each hole is around 5~mm.
It is worth noting that cushions are attached to the bottom of the plates and pegs to reduce the load when the insertion attempts to deviate from the center of the holes.
In this setting, the working space of the end effector is within $[\underline{p}, \overline{p}]$ (see Table~\ref{tab:exp_environment}).

The three-dimensional observed and commanded positions of the end effector are used as RL state space.
To utilize the load as collision detection, the observed currents on four axes (without the actuator for the gripper) are added (roughly within [$-$500~mA, 500~mA] for each current).
From the control error and the load, the flag of collision ($\in [0, 1]$) is stochastically given.
In total, the 11-dimensional state space is constructed.

The action space is defined as three-dimensional: the velocity of the end effector in Cartesian space is integrated into the commanded position using the Euler method.
The velocity of the end effector is bounded within 0.01, 0.015, and 0.005~m/s for $x$-, $y$-, $z-$axes, respectively.
In addition, the commanded position is also clipped within $[\underline{p}, \overline{p}]$.
In the control loop of the robot (with around 3~Hz), as the probability of the collision is increased, proportional feedback control is activated so that the commanded position returns to the collision point (i.e., the current observed position).

To create a reward function, we define the task as reaching the target position in the corresponding hole while minimizing load as much as possible.
This purpose is implemented as the following reward function:
\begin{align}
    r = \exp(- k \|p^\mathrm{obs} - p^\mathrm{tar}\|_\Sigma^2) - f_c
    + \begin{cases}
        \pm 100 & p^\mathrm{obs}_z \leq p^\mathrm{tar}_z
        \\
        0 & \mathrm{otherwise}
    \end{cases}
\end{align}
where $p^\mathrm{obs}$ denotes the observed position of the end effector, $p^\mathrm{tar}$ is the target position (i.e., the rough position of the center of the hole).
$\Sigma$ is for scaling the axes by the ranges of the commanded position $[\underline{p}, \overline{p}]$, and $k$ denotes the gain for sparsity.
$f_c$ denotes the stochastic flag of collision.
If the peg is inserted into the correct hole, the successful bonus $100$ is given; and if the peg is inserted into the wrong hole, the big penalty $-100$ is given.
Whether the inserted hole is correct or not is roughly judged according to $(p^\mathrm{obs}_y  p^\mathrm{tar}_y) > 0$.
In both cases, that episode is terminated and a new episode is started after initialization.
It is worth noting that the first term for approaching the target is auxiliary because the precise target position cannot be specified in an environment that has not been precisely built, whereas only the last sparse term would make the task intractable.

To demonstrate the effectiveness and adaptability of the eligibility traces, the following demonstration is performed.
Here, it is noticed that the proposed eligibility traces with $(\lambda_\mathrm{max}^1, \lambda_\mathrm{max}^2, \kappa) = (0.5, 0.9, 1.0)$ outperformed the conventional ones in the above simulations, only the proposed method represents the eligibility traces.
In the first stage, the robot attempts to insert a peg with a flower shape into the hole on the left side, and the methods with and without eligibility traces are compared to demonstrate the sample efficiency of the eligibility traces even in a real robot experiment.
Following the first stage, the peg is replaced with one with a pentagon shape for the hole on the right side, and the robot moves on to the second stage of learning the peg-in-hole task with a different target.
Because the robot can only detect this change through reward signals, the robot's previous raw experiences at the first stage would prevent it from adapting to the new target.
Even in that case, the eligibility traces would efficiently adapt the robot to the target change due to no reuse of previous raw experiences.
The above configurations are summarized in Table~\ref{tab:exp_environment}.
To accomplish this task, the proposed method with the same parameters as Table~\ref{tab:sim_parameter} except for the learning rate is employed.
Since the real robot experiment would contain large uncertainty due to observation and control noises, the learning rate $\alpha$ is reduced to $5\times10^{-5}$ for stable learning.
The length of episodes in each stage is differently given: 200 episodes for the first stage and; 100 episodes for the second stage.

\begin{table*}[tb]
    \caption{Configurations for peg-in-hole task}
    \label{tab:exp_environment}
    \centering
    \begin{tabular}{ccc}
        \hline\hline
        Symbol & Meaning & Value \\
        \hline
        $|\mathcal{S}|$ & State space & 11 \\
        $|\mathcal{A}|$ & Action space & 3 \\
        $T$ & Maximum time step & 180 ($\sim$60 s) \\
        $E$ & Number of episodes & (200, 100) \\
        $\underline{p}$ & Lower bound of position & (0.09~m, -0.07~m, 0.024~m) \\
        $\overline{p}$ & Upper bound of position & (0.16~m, 0.07~m, 0.044~m) \\
        $p^\mathrm{ini}$ & Initial position & (0.1~m, 0.0~m, 0.04~m) \\
        $p^\mathrm{tar}$ & Goal position & (0.125~m, $\pm$0.035~m, 0.025~m) \\
        $k$ & Gain for reward scaling & 5.0 \\
        \hline\hline
    \end{tabular}
\end{table*}

\begin{figure*}[tb]
    \centering
    \includegraphics[keepaspectratio=true,width=0.98\linewidth]{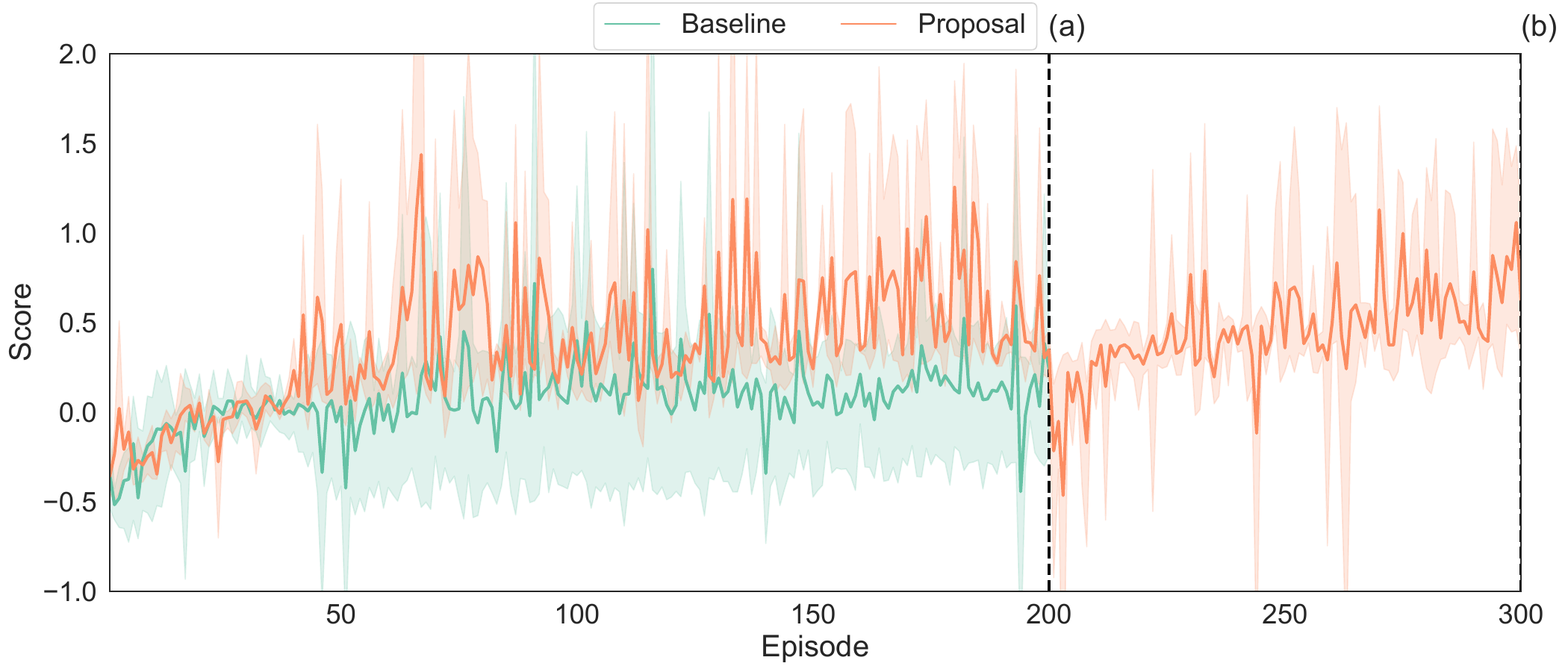}
    \caption{Learning curves by DRLs without/with the proposed eligibility traces (Baseline and Proposal, respectively):
        score in the vertical axis denotes the mean of the rewards obtained in each episode;
        after 200 episodes, the peg grasped was exchanged for the new target;
        the proposal accelerated learning speed and yielded the peg insertion, while the baseline hardly achieved the task;
        even after exchanging the peg grasped and the target, the proposal properly adapted the robot to that difference by fully utilizing the streaming data.
    }
    \label{fig:exp_learn}
\end{figure*}

\begin{figure*}[tb]
    \centering
    \subfigure[At the end of the first stage]{
        \includegraphics[keepaspectratio=true,width=0.48\linewidth]{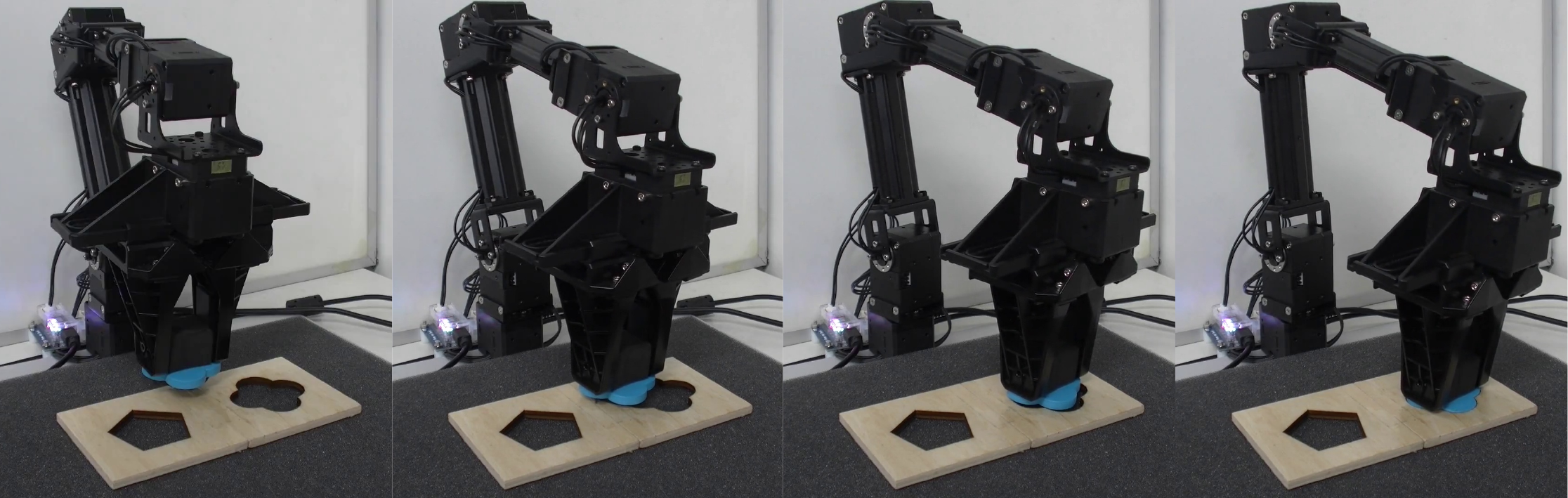}
    }
    \centering
    \subfigure[At the end of the second stage]{
        \includegraphics[keepaspectratio=true,width=0.48\linewidth]{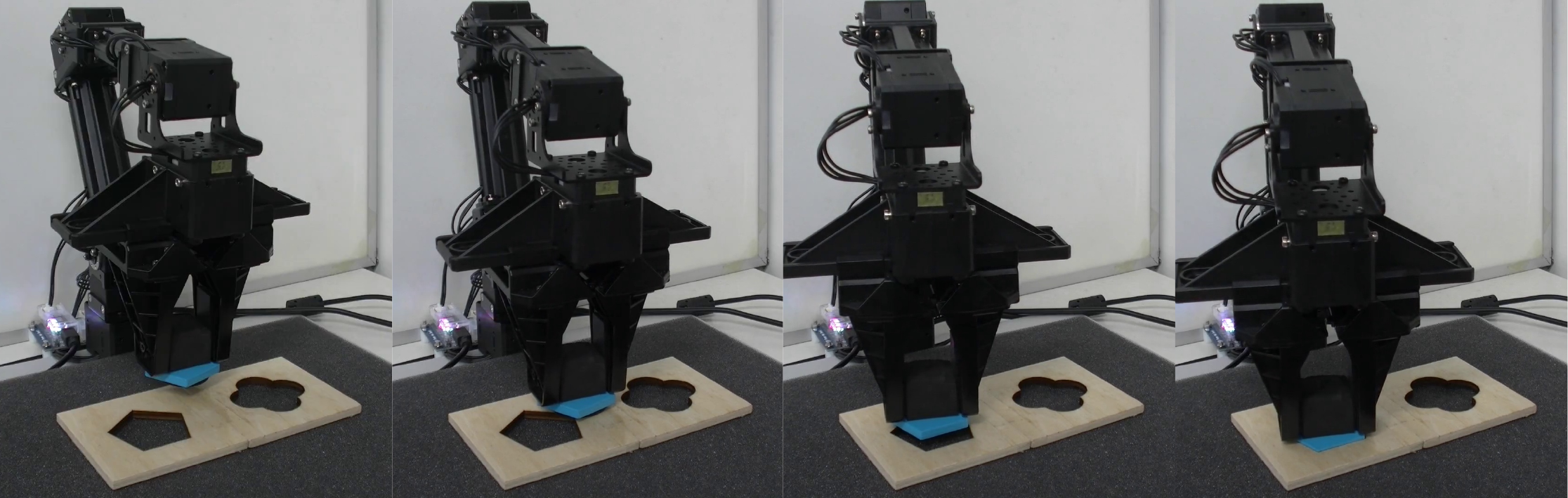}
    }
    \caption{Snapshots at the times pointed in Fig.~\ref{fig:exp_learn}:
        in both stages, the robot succeeded in inserting the peg into the corresponding hole.
    }
    \label{fig:exp_snap}
\end{figure*}

\subsection{Results}

The learning curves from three trials are depicted in Fig.~\ref{fig:exp_learn}.
The successful snapshots of the test results at the times pointed in Fig.~\ref{fig:exp_learn} (i.e., the ends of the respective stages) are also shown in Fig.~\ref{fig:exp_snap}.

During the first stage, the robot had the peg corresponding to the hole with a flower shape on the left side of the robot.
As can be seen in Fig.~\ref{fig:exp_learn}, the case without the proposed eligibility traces was mostly trapped into local optima, and could not insert the peg into the corresponding hole.
In contrast, the proposed eligibility traces accelerated learning speed and resulted in the peg insertion numerous times during learning.
After finishing the first stage (at 200 episodes), the robot succeeded in demonstrating the acquired peg-in-hole skill as shown in Fig.~\ref{fig:exp_snap}(a) (also see the attached video).

Following the first stage, the peg grasped by the robot was replaced with the one corresponding to the hole with a pentagon shape on the robot's right side.
Because the robot was unaware of this fact, at the start of the second stage, it failed to approach the correct hole and received a large penalty for inserting the peg grasped into the incorrect hole.
However, after these failures, the robot began to search for the new target, which it discovered on the opposite side.
The robot could rapidly adapt to the new target by fully utilizing the streaming data without storing the previous experiences because it knew how to insert the peg into the hole (i.e., the pushing down motion has a large reward).

\subsection{Discussion}

Consequently, we demonstrated the efficacy of the eligibility traces method and its enhancement through the proposed adaptive and multiple time-scale eligibility traces.
However, when compared to the simulation results, the learning curves of the real robot experiments are not yet stable, implying the need to develop new learning methods and regularization techniques that are robust to noise in the real environment.
Robustness is especially important when using eligibility traces because stable learning gradients derived from the average of a large amount of data cannot be used.
However, we must exercise caution in adjusting the tradeoff between conservative learning and sample efficiency so that we do not reduce sample efficiency for the sake of conservative learning.

Furthermore, in terms of RL theory, eligibility traces can be integrated not only with the actor-critic-based algorithms used in the preceding verification, but also with other algorithms, such as those suitable for discrete action spaces, such as SARSA.
We used the actor-critic algorithm, which can handle continuous action space, in this paper because it is assumed to be used for robot control, but we should also evaluate its performance when combined with other algorithms.
It is worth noting that, just as traditional eligibility traces were effective for RL with linear regressors, the proposed method for conservatively alleviating the problem caused by nonlinear regressors is expected to be effective for any DRL.

\section{Conclusion}

This paper proposed new eligibility traces with adaptive decaying and multiple time-scale traces for sample-efficient DRL without storing raw experiences.
The gradient divergence was proposed as the reason why the DRL with the eligibility traces fails to learn.
As a result, when the output divergence (rather than the gradient divergence in the pursuit of low computational cost) is increased, the accumulated traces are reset by adaptively adjusting the decaying factor.
Furthermore, the replacing operation in the replacing eligibility traces would mitigate the negative effects of gradient divergence, though gradient-by-gradient replacing cannot improve sample efficiency.
To introduce this capability with high sample efficiency, a new update rule with generalized eligibility traces, including standard and replacing ones, was designed.
By replacing operation, it stores the most dominant short-term traces with sufficient sample efficiency into long-term traces.
Four types of simulations were first performed to validate the proposed method in terms of general learning performance.
In terms of the sum of rewards obtained by the trained policies and sample efficiency, the generalized eligibility traces outperformed the conventional versions.
While the strengths of the standard and replacing eligibility traces varied, the proposed method allowed the agent to solve all tasks stably by inheriting their characteristics.
The peg-in-hole demonstration with a robot arm was also shown to demonstrate adaptability to non-stationary problems.
We found that even if the target (i.e., the reward function) changed over time, the robot could still achieve the new target by efficiently following the target change.
Future work is to theoretically investigate the optimal design of the generalized eligibility traces, even with any RL algorithms like TD($\lambda$).
In addition, from the perspective of real-world robotic applications, even if the nominal dynamics are known, changes from that (due to individual differences, aging, etc.) often make significant adverse effects on performance, although the demonstration in this paper dealt with changes in reward.
In order to deal with such problems while making the best use of prior knowledge of robotics such as nominal dynamics, it would be desirable to appropriately integrate the model-based RL~\cite{thuruthel2018model,clavera2020model} and residual learning~\cite{johannink2019residual,kulkarni2021learning} while carefully considering catastrophic forgetting of the model learned~\cite{kobayashi2020reinforcement}, instead of theoretically improving the proposed method.
After establishing the sample-efficient online DRL, we will apply it to more complicated real-world applications with autonomous robots.

\section*{Acknowledgments}

This work was supported by JSPS KAKENHI, Grant-in-Aid for Scientific Research (B), Grant Number JP20H04265.

%
%
\section*{\refname}
\bibliographystyle{elsarticle-num}
\bibliography{biblio}

\end{document}